\documentclass{article}

\PassOptionsToPackage{numbers}{natbib}

\usepackage[final]{nips_2017}

\usepackage[utf8]{inputenc} 
\usepackage[T1]{fontenc}    
\usepackage{hyperref}       
\usepackage{url}            
\usepackage{booktabs}       
\usepackage{amsfonts}       
\usepackage{nicefrac}       
\usepackage{microtype}      
\usepackage{graphicx}
\usepackage[ruled]{algorithm2e}
\usepackage[title]{appendix}
\SetKwRepeat{Do}{do}{while}%

\usepackage{amsmath,amssymb,mathtools,bm,etoolbox}

\DeclareMathOperator*{\E}{\mathbb{E}}

\title{Improvements on Hindsight Learning}

\author{
  Ameet Deshpande \\
  Department of CSE\\
  Indian Institute of Technology Madras\\
  \texttt{ameetsd97@gmail.com} \\
  \And
  Srikanth Sarma \\
  Department of ME \\
  Indian Institute of Technology Madras \\
  \texttt{gsarma8@gmail.com} \\
    \And
  Ashutosh Jha \\
  Department of ME \\
  Indian Institute of Technology Madras \\
  \texttt{ashutosh1296@gmail.com} \\
    \And
  Balaraman Ravindran \\
  Department of CSE \\
  Indian Institute of Technology Madras \\
  \texttt{ravi@cse.iitm.ac.in} \\
}

\begin{document}

\maketitle

\begin{abstract}
Sparse reward problems are one of the biggest challenges in Reinforcement Learning. Goal-directed tasks are one such sparse reward problems where a reward signal is received only when the goal is reached. One promising way to train an agent to perform goal-directed tasks is to use Hindsight Learning approaches. In these approaches, even when an agent fails to reach the desired goal, the agent learns to reach the goal it achieved instead. Doing this over multiple trajectories while generalizing the policy learned from the achieved goals, the agent learns a goal conditioned policy to reach any goal. One such approach is Hindsight Experience replay which uses an off-policy Reinforcement Learning algorithm to learn a goal conditioned policy. In this approach, a replay of the past transitions happens in a uniformly random fashion. Another approach is to use a Hindsight version of the policy gradients to directly learn a policy. In this work, we discuss different ways to replay past transitions to improve learning in hindsight experience replay focusing on prioritized variants in particular. Also, we implement the Hindsight Policy gradient methods to robotic tasks.
\end{abstract}

\section{Introduction}

Sparse reward settings are useful because the reward function engineering is easy. A simple $+1$ for achieving the required goal and $0$ otherwise is one example. Recent work on Hindsight Learning \cite{HER} has shown that learning from trajectories in which the agent does not succeed in achieving the goal can improve performance greatly. This alleviates the sparse reward problem by ensuring there are transitions with non-zero rewards in \textit{every} rollout. A replay buffer is used to improve the sample efficiency of procedure \cite{mnih2015human}. 

Prioritized Experience Replay \cite{PER}, a variant of Experience Replay, samples the transitions based on priority values assigned to them, unlike uniform sampling that is followed in vanilla Experience Replay. Though in principle the priority values could be calculated using any formulation, it is common to use the TD error as a proxy. This follows from the intuition that a large TD error indicates a shortcoming of the agent in the learning about that part of the environment.

In this paper, we propose \textit{Hindsight Prioritized Experience Replay}, a variant which aims to leverage the best of both the worlds. It can be used in settings which have multiple goals, which is common in robotic tasks.

\section{Background and Notation}
This section introduces the concepts motivating Hindsight Learning and Prioritized Experience Replay.

\subsection{Hindsight Experience Replay}
Humans seem to learn both from successes and failures. In sparse reward settings, the agent does not gather much information from trajectories in which it failed to achieve the goal because the total return ($G_t$) for all the steps is $0$. Consider the example of an agent being trained to play soccer to understand why it could benefit from learning from its mistakes.

\begin{figure}[h]
  \centering
	\includegraphics[width=.6\textwidth]{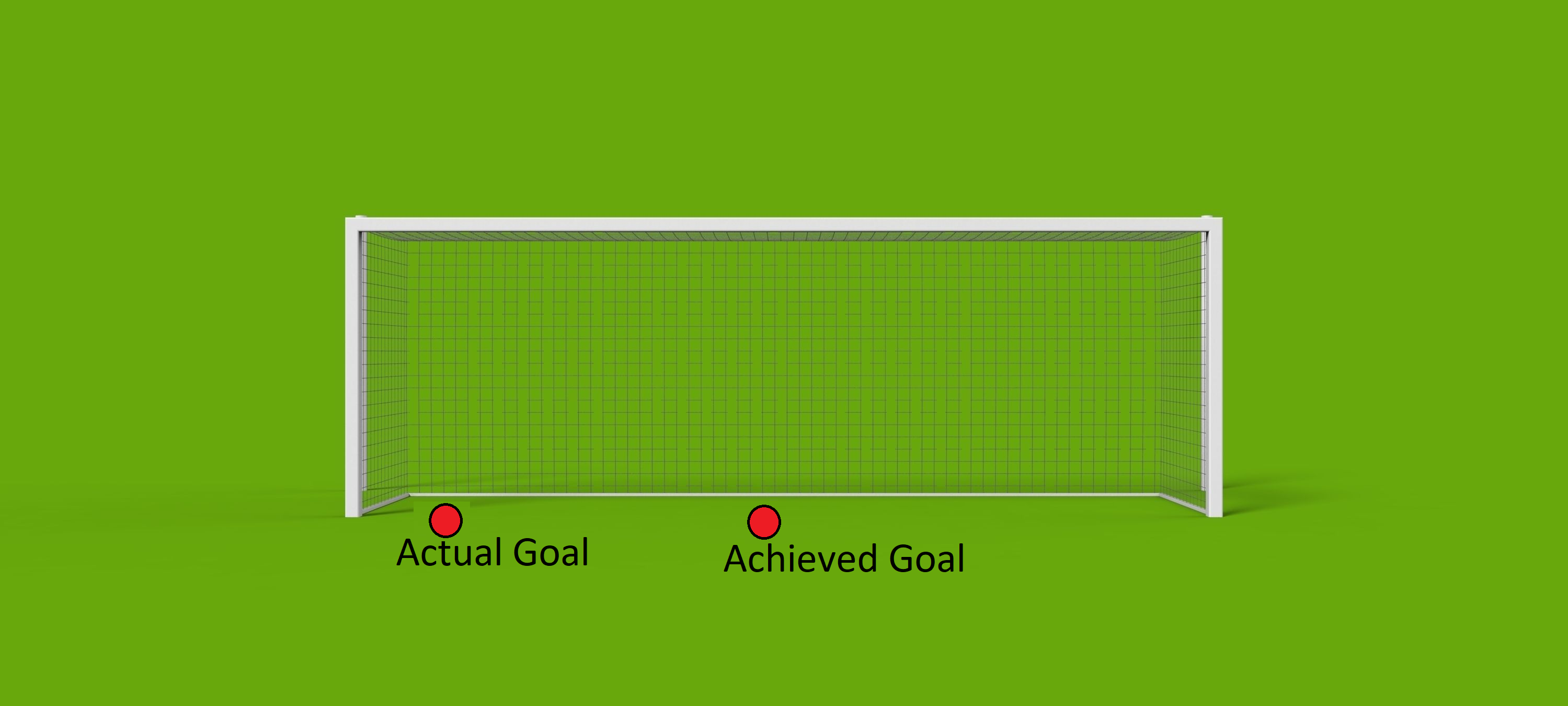}
\end{figure}


Though the agent is not able to achieve its goal, it can take away important information about how it needs to kick the ball if at all it was told to kick to the center of the goal post.

This work is based on the idea of Universal Value Function Approximators \cite{schaul2015universal} where the reward function is conditioned on the goal, $r_g(s,a)$. A goal $g$ is chosen at the start of the episode and remains the same throughout. It has been shown that this approach can be used to generalize to previously unseen state-goal-action pairs.

The strategy for learning from alternate goals is called \texttt{future}. We refer a reader seeking a detailed explanation of the same to \cite{HER}, but the key idea is to consider future states $s'$ of the episode, with respect to a state $s_t$ and use that as an alternate goal. Consider the following episode where the agent failed to achieve the required goal.
\[ s_1 \to s_2 \to \dots \to s_k \to \dots \to s_{T} \] If the state $s_k$ is considered as an alternate goal, a reward of $+1$ is used to backup the values of all the states before it. Using this intuitive strategy, the following is the pseudo code of the algorithm.

\begin{algorithm}[H]
\SetAlgoLined

\Do{$continue\_learning$}{
\For{episode in max\_episodes}{Store trajectory in buffer $D$}

Sample $batch$ number of transitions \;
Set $alternate=\frac{replay\_k}{1+replay\_k}$\;
For $alternate$ fraction of transitions, pick a future state at random\;
Use this modified sample to learn\;
}

\caption{Hindsight Experience Replay}
\end{algorithm}

A key thing to note in the algorithm is that after the agent experiences an episode, it just stores it in the buffer and the sampling of future states in the \texttt{future} strategy happens at learning time. This ensures that in expectation, any future state of a state $s_t$ has an equal probability of getting picked. We refer to this as \textit{uniform sampling} and this plays a key role in our experiments.

\subsection{Prioritized Experience Replay}

A work on prioritizing transitions \cite{PER} showed that prioritizing transitions based on their TD-error serves as a good proxy for the amount of learning that can happen with it. An ideal implementation would involve sorting all the transitions based on their TD-error after each rollout, but this is infeasible in practice. There are two strategies considered for serving as an approximation for this, \texttt{proportional} and \texttt{rank-based}. Interested readers are referred to \cite{PER} for more details, but it was observed that \texttt{rank-based} performed better.

This strategy uses a priority queue for storing the transitions, with the TD-error $\delta$ as the key value. This reduces the insertion time to $log(N)$, where $N$ is the size of the priority queue.

\subsection{Hindsight Policy Gradients}
Hindsight Experience relies on an off-policy Reinforcement Learning algorithm as it uses a replay buffer to de-correlate the training samples. One other way to de-correlate training samples is to use Asynchronous methods as mentioned in \cite{A3C}. The best performing algorithm based on Asynchronous methods, Asynchronous Advantage Actor Critic algorithm A3C learns both a value function and a policy. To learn a policy, one must compute the policy gradients i.e, the gradients of the average expected reward with respect to the policy parameters. In a recent work \cite{HPG}, policy gradients were derived to incorporate hindsight learning for the case of goal directed tasks. The expression for the policy gradient as mentioned in \cite{HPG} is
\begin{equation}
\delta_{HPG} = \sum_g p(g) \left[\prod_{t=1}^{T-1}\frac{\pi(a_t|s_t,g; \theta)}{\pi(a_t|s_t,g'; \theta)} \right] \sum_{t=1}^{T-1} \nabla \log{\pi(a_t | s_t, g; \theta)} \sum_{t'=t+1}^{T} r(s_{t'}, g)
\end{equation}
 where $g'$ is the goal that is being pursued, $g$ are the goals that were achieved in the process, $\pi$ is the policy parametrized by parameters $\theta$. It has been shown in \cite{HPG} that $\delta_{HPG}$ can be used to train agents to perform goal directed tasks. But, the tasks have were limited to bit flipping and empty grid. This approach can directly be extended to continuous state and action spaces. In addition, the policy gradient estimate $\delta_{HPG}$ can be improved by using baseline corrections which paves a way for A3C.
The policy gradient in the case of A3C would be
\begin{equation}
\delta_{HPG} = \sum_g p(g) \left[\prod_{t=1}^{T-1}\frac{\pi(a_t|s_t,g; \theta)}{\pi(a_t|s_t,g'; \theta)} \right] \sum_{t=1}^{T-1} \nabla \log{\pi(a_t | s_t, g; \theta)} \sum_{t'=t+1}^{T} \left(r(s_{t'}, g) - V^{\pi}(s_{t'} || g)\right)
\end{equation}
 
\section{Hindsight Prioritized Experience replay}

The intuition for this algorithm is to have the best of both worlds. The vanilla replay buffer is replaced with a priority queue to implement the \texttt{rank-based} strategy. There is one major difference though. The fact that we are using a priority-queue means that \textit{we necessarily have to} to calculate the TD-error before storing. This means that the goals $g$ will be required to calculate the reward using the conditional function ($r_{g}(s,ac$) that we discussed in Section $2$. This forces us to sample the alternate goals \textit{before} storing.

\subsection{Controlling the ratio of Actual goals and Alternate goals}

In the original work \cite{HER}, $replay\_k$ ensures that exactly $\frac{1}{replay\_k+1}$ fraction of sampled transitions are from actual goals and $\frac{replay\_k}{replay\_k+1}$ fraction of sampled transitions are for alternate goals. In the modified method, since we are storing the transitions based on their TD-error and then sampling, there is no restriction on the ratio of actual goals and alternate goals in the sample. A method called \texttt{two\_queues} is developed to ensure the same. The idea is very simple and uses $2$ priority queues instead of $1$.

\begin{enumerate}
\item Instantiate two queues
\item Observe a trajectory
\item For each transition choose $n_{alt}$ number of alternate goals
\item Push the transitions with actual goals to priority queue $1$
\item Push the transitions with alternate goals to priority queue $2$
\item At sample time, keep the ratio of sampling from the two priority queues as $1:replay\_k$
\end{enumerate}

\subsection{Caveats of uniform sampling}

One caveat both in the method developed so far and in \cite{HER} is the uniform sampling of goals that was discussed in Section $2.1$. Sampling alternate goals is done by uniformly sampling from \texttt{future} states. Each state in the trajectory has equal chance to get picked. Consider a trajectory $s_1 \to s_2 \to s_3 \to s_4 \to s_5$. If all the states have an equal probability to get picked, the probability of picking the state goal pair $s=s_4,g=s_5$ is $4$ times more than the probability of picking $s=s_1,g=s_5$. This is because there are $4$ different options for $s_1$, whereas there is only $1$ for $s_4$. This effect will be observed in the tail of all the trajectories.

This problem has worse effects on our method because it stores $n_{alt}$ number of alternate goals for each transition. There are thus multiple copies of the same transition in the buffer and updating the values of one of them does not change the other. To alleviate this, we propose a method called \texttt{non-uniform sampling}. Instead of choosing $n_{alt}$ number of goals per transition, if the transition is observed at time $t$ with episode length $T$, choose a number characterized by the formula below.\[ \frac{t}{T}\times n_{alt} \]

As shown in the experiments later, this method gave the best performance.

\subsection{Annealing replay\_k}

With the ratio $\frac{n_{act}}{n_{alt}}$ set to $\frac{1}{replay\_k}$, the agent is stuck with this throughout the training procedure. A very high value of $replay\_$ means that the agent cares only about alternate goals, and that is not desirable. However, a low value of $replay\_k$ means that Hindsight Learning is not being leveraged. Like other trade-offs, hyperparameter tuning over $replay\_k$ is required. But having a constant $replay\_k$ throughout learning may not be ideal. Consider this graph which has a very high variance.

\begin{figure}[h]
    \centering
    \includegraphics[width=.6\textwidth]{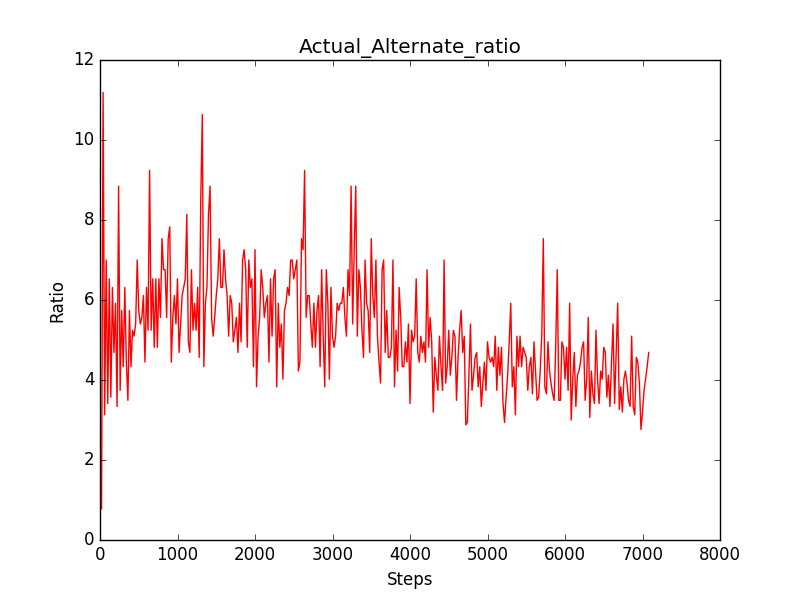}
    \caption{Actual Alternate Ratio}
    \label{fig:aar}
\end{figure}

The ratio seems to be falling slowly on average from about $6$ to $4$. In the initial stages of learning when the agent makes more mistakes, it might make sense for it to learn more from its mistakes and as it becomes more experienced, it might care only about the actual goals. There are two ways to give the agent this freedom.

\subsection{Using a single priority queue}
The second solution discussed in section $3.3$ is implemented. Instead of maintaining two different queues, only $1$ queue is maintained. But to ensure that the ratio of actual to alternate goals is such that actual goals are still being used to learn, we set $n_{alt}=replay\_k$. In the case when the TD-error of all the transitions is the same, the ratio of actual goals and alternate goals is $\frac{1}{replay\_k}$. This method is referred to as \texttt{single\_queue}.

\begin{enumerate}
\item Anneal $replay\_k$ by following some schedule. There are popular schedules like the ones used for learning rates \cite{darken1991note}
\item Use a single queue instead of \texttt{two\_queues} and hence allow the TD-error to decide the ratio. This was the scheme used to generate Figure \ref{fig:aar}.
\end{enumerate}

\subsection{Algorithm}

The following is the pseudo code for the algorithm. The main difference from Algorithm $1$ is that goals are sampled at storage time.
\begin{algorithm}[H]
\SetAlgoLined

\Do{$continue\_learning$}{
\For{episode in $max\_episodes$}{Observe trajectory $T$\;
\For{$s_{t},a_{t},s_{t+1}$ in $T$}{
$num\_goals \gets \left(1-\frac{t}{T}\right)\times replay\_k$\;
Sample $num\_goals$ number of goals \;
append transition with sampled goals and actual goal \;
Compute priority $\delta$ for each goal appended transition \;
$\delta = r + \gamma Q(S_{t+1} || g, a_{t+1}) - Q(S_{t} || g, a_{t})$ \;
Push goal appended transition to priority queue $P_{1}\ (P_{2})$\;
}

Sample \textit{batch} number of transitions\;
Use sample to learn\;
}

\caption{Hindsight Prioritized Experience Replay}
}
\end{algorithm}

\section{Experiments}

The following results illustrate the experiments that were performed. A more thorough analysis is deferred to the appendix.


\begin{figure}[h]
    \centering
    \begin{minipage}{0.4\textwidth}
    \centering
    \includegraphics[width=\textwidth]{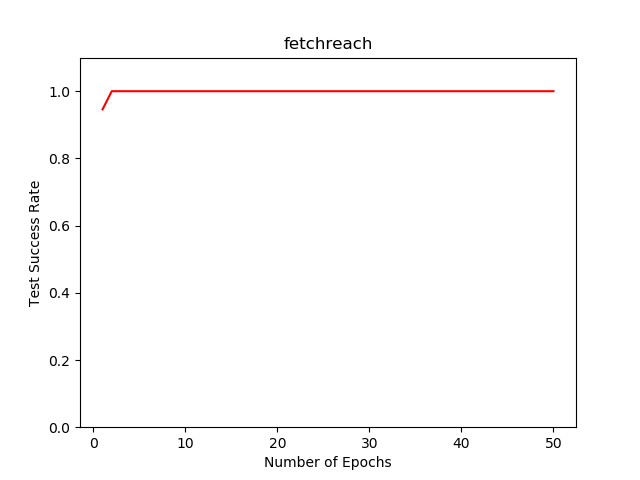}
    \end{minipage}
    \begin{minipage}{0.4\textwidth}
    \centering
    \includegraphics[width=\textwidth]{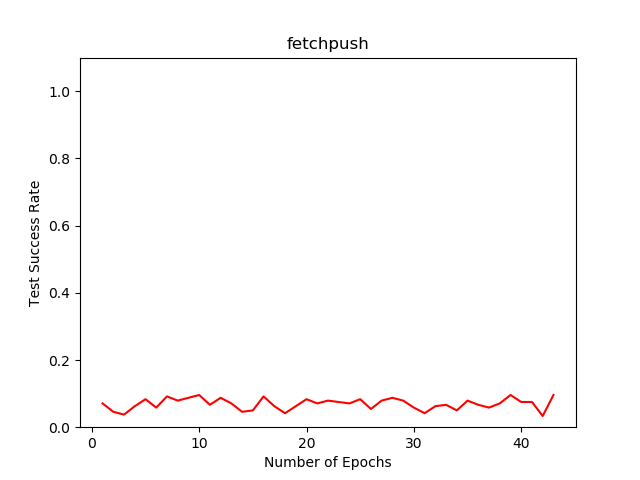}
    \end{minipage}
    \centering
    \begin{minipage}{0.4\textwidth}
    \centering
    \includegraphics[width=\textwidth]{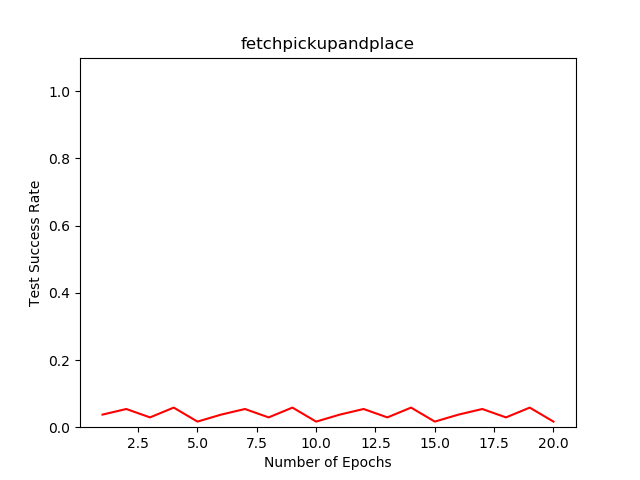}
    \end{minipage}
    \begin{minipage}{0.4\textwidth}
    \centering
    \includegraphics[width=\textwidth]{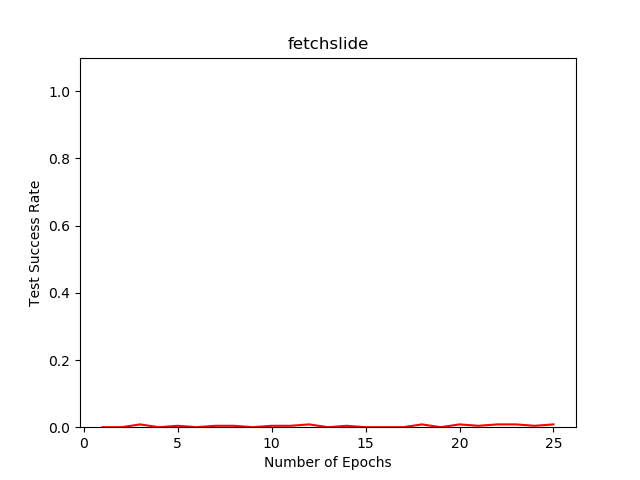}
    \end{minipage}
    \caption{Fetch Environments}
\end{figure}

There are oscillations that are visible in the plots. The oscillations decreased only marginally when the batch size was increased to $512$ from $256$. The proposed algorithm reaches the optimal performance for the \textit{FetchReach} environments in the same amount of training time as \cite{HER}, but it fails on more complex environments.

\begin{tabular}{||c|c|c|c|c|c|c||}
\hline
Environment Name &	$replay\_k$  &	$batch\_size$ &	$n\_batches$ &	Best success rate  &	Best epoch &	Strategy\\
\hline
\hline
FetchReach-v1 &	4 & 512 &	50 &	1 &	1 &	$single\_queue$\\
\hline
FetchReach-v1 &	6 &	512 &	50 &	1 &	1 &	$single\_queue$\\
\hline
FetchSlide-v1 &	4 &	512 &	50 &	0.0167 &	26 &	$single\_queue$\\
\hline
FetchPush-v1 &	8 &	256 &	50 &	0.09583 &	38 &	$single\_queue$\\
\hline
FetchPickAndPlace-v1 &	8 &	256 &	50 &	0.05833 &	18 &	$single\_queue$\\
\hline
FetchSlide-v1 &	8 &	512 &	50 &	0.12083 &	20 &	$single\_queue$\\
\hline
\hline
FetchReach-v1 &	6 &	512 &	40 &	1 &	1 &	$two\_queues$\\
\hline
FetchReach-v1 &	8 &	512 &	40 &	1 &	1 &	$two\_queues$\\
\hline
FetchPush-v1 &	4 &	256 &	40 &	0.09583 &	16 &	$two\_queues$\\
\hline
FetchPush-v1 &	6 &	256 &	40 &	0.1125 & 45 &	$two\_queues$\\
\hline
FetchPush-v1 &	8 &	256 &	40 &	0.083 &	47 &	$two\_queues$\\
\hline
FetchPush-v1 &	8 &	512 &	40 &	0.09583 &	1 &	$two\_queues$\\
\hline
\end{tabular}

\section{Future Work}
The results of Prioritized Experience Replay were unsatisfactory. A major shortcoming of our approach is the fact that the method is tied to calculate the TD-error before storing, while the other solution is computationally intractable. Though prioritized experience replay used TD-error as a proxy, to the best of the authors' knowledge, there is no method which uses some heuristics/algorithm to suggest which transition may be more important. Some example heuristics could be, ``If a state-goal pair has high TD-error, other goals with the same states have a high error as well'', ``Some goals are hard to achieve for a large number of states'', and so on. An ideal case would be to realize a mode which can suggest what transitions to train on \cite{plappert2018multi}. This gives a flavor of Model-Based Learning and Prioritized Sweeping and this will be the focus of the authors in the future. The current interest in the community to make prioritized methods more efficient \cite{horgan2018distributed} leaves a lot of promise.

The initial results of Hindsight policy gradients in bit flipping and empty grid environments show good promise. This leaves one to extend the hindsight policy gradients beyond vanilla policy gradient methods to more advanced methods like A2C/ A3C and TRPO/ PPO.

\section{Conclusion}
With efficient and faster ways of performing comes more data, but a uniform sampling of this data may not be the best way to harness it. Reinforcement Learning setups have an inherent way of assigning weights to states based on their importance (state visitation frequencies on successful trajectories). If the TD-error calculation becomes a bottleneck, most of the computational resources should be spent on important transitions. Prioritized Experience Replay offers one such method, but as analyzed in the paper, it does not work very well. Unless there is a way to compute the TD error after sampling and yet have a priority assigned to a transition while storing, the authors don't see much hope in the traditional methods.

Online versions of Hindsight Experience Replay can benefit from multiple optimization techniques that have been developed for the same. But at the moment, the variance reduction techniques that have been used for Hindsight Policy Gradients are not sufficient to get complex environments working. Performance even in simple environments like \textit{bit-flipping} is also only satisfactory. There is a need to come up with a more principled approach to multi-goal RL.




\bibliographystyle{plain}
\bibliography{ref}

\begin{thebibliography}{10}

\bibitem{HER}
Marcin Andrychowicz, Filip Wolski, Alex Ray, Jonas Schneider, Rachel Fong,
  Peter Welinder, Bob McGrew, Josh Tobin, OpenAI~Pieter Abbeel, and Wojciech
  Zaremba.
\newblock Hindsight experience replay.
\newblock In {\em Advances in Neural Information Processing Systems}, pages
  5048--5058, 2017.

\bibitem{darken1991note}
Christian Darken and John~E Moody.
\newblock Note on learning rate schedules for stochastic optimization.
\newblock In {\em Advances in neural information processing systems}, pages
  832--838, 1991.

\bibitem{horgan2018distributed}
Dan Horgan, John Quan, David Budden, Gabriel Barth-Maron, Matteo Hessel, Hado
  van Hasselt, and David Silver.
\newblock Distributed prioritized experience replay.
\newblock {\em arXiv preprint arXiv:1803.00933}, 2018.

\bibitem{ddpg_per}
Yuenan Hou, Lifeng Liu, Qing Wei, Xudong Xu, and Chunlin Chen.
\newblock A novel ddpg method with prioritized experience replay.
\newblock In {\em Systems, Man, and Cybernetics (SMC), 2017 IEEE International
  Conference on}, pages 316--321. IEEE, 2017.

\bibitem{A3C}
Volodymyr Mnih, Adria~Puigdomenech Badia, Mehdi Mirza, Alex Graves, Timothy
  Lillicrap, Tim Harley, David Silver, and Koray Kavukcuoglu.
\newblock Asynchronous methods for deep reinforcement learning.
\newblock In {\em International Conference on Machine Learning}, pages
  1928--1937, 2016.

\bibitem{mnih2015human}
Volodymyr Mnih, Koray Kavukcuoglu, David Silver, Andrei~A Rusu, Joel Veness,
  Marc~G Bellemare, Alex Graves, Martin Riedmiller, Andreas~K Fidjeland, Georg
  Ostrovski, et~al.
\newblock Human-level control through deep reinforcement learning.
\newblock {\em Nature}, 518(7540):529, 2015.

\bibitem{plappert2018multi}
Matthias Plappert, Marcin Andrychowicz, Alex Ray, Bob McGrew, Bowen Baker,
  Glenn Powell, Jonas Schneider, Josh Tobin, Maciek Chociej, Peter Welinder,
  et~al.
\newblock Multi-goal reinforcement learning: Challenging robotics environments
  and request for research.
\newblock {\em arXiv preprint arXiv:1802.09464}, 2018.

\bibitem{HPG}
Paulo Rauber, Filipe Mutz, and Juergen Schmidhuber.
\newblock Hindsight policy gradients.
\newblock {\em arXiv preprint arXiv:1711.06006}, 2017.

\bibitem{schaul2015universal}
Tom Schaul, Daniel Horgan, Karol Gregor, and David Silver.
\newblock Universal value function approximators.
\newblock In {\em International Conference on Machine Learning}, pages
  1312--1320, 2015.

\bibitem{PER}
Tom Schaul, John Quan, Ioannis Antonoglou, and David Silver.
\newblock Prioritized experience replay.
\newblock {\em arXiv preprint arXiv:1511.05952}, 2015.

\bibitem{TRPO}
John Schulman, Sergey Levine, Pieter Abbeel, Michael Jordan, and Philipp
  Moritz.
\newblock Trust region policy optimization.
\newblock In {\em International Conference on Machine Learning}, pages
  1889--1897, 2015.

\end{thebibliography}



\newpage
\begin{appendix}

\section{Bias correction for PER sampling}
Prioritized experience replay induces a bias as it changes the distribution of the sample transitions used for updates. The loss minimization for learning the value function relies on the sampling being uniform. Hence, to make the updates unbiased, we must introduce an importance sample ratio given by $1/(N*P(i))$ where P(i) is the probability of picking that transition whereas N is the size of the buffer. In order to make the updates unbiased towards the end of training while still exploiting the benefits of prioritized sampling, we anneal the bias by using an importance sample of the form $(1/(N*P(i)))^\beta$. The value of $\beta$ is annealed over time such that towards the end of training, $\beta \to 1$. As mentioned in \cite{ddpg_per}, bias correction is used only for the value function updates and not for the policy updates. This showed better performance than using bias correction for both value function and policy updates.

\section{Evaluating Baselines}

The baseline code provided by \cite{HER} is evaluated on all the MuJoCo environments. The hyperparameters provided by their implementation were used along with $num\_cpu=12$, which means that the code is run on $12$ CPUs. Though they recommend usage of $19$ CPUs, we had to stick to a lower number because of unavailability of computational resources. The following plots illustrate the success rates.

\begin{figure}[h]
    \centering
    \begin{minipage}{0.4\textwidth}
    \centering
    \includegraphics[width=\textwidth]{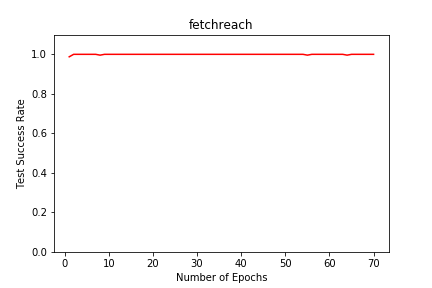}
    \end{minipage}
    \begin{minipage}{0.4\textwidth}
    \centering
    \includegraphics[width=\textwidth]{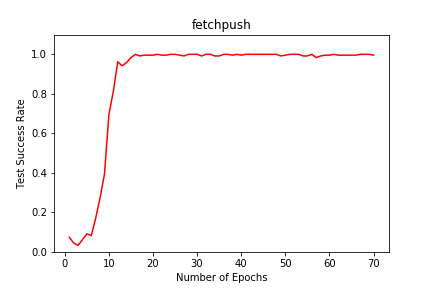}
    \end{minipage}
    \centering
    \begin{minipage}{0.4\textwidth}
    \centering
    \includegraphics[width=\textwidth]{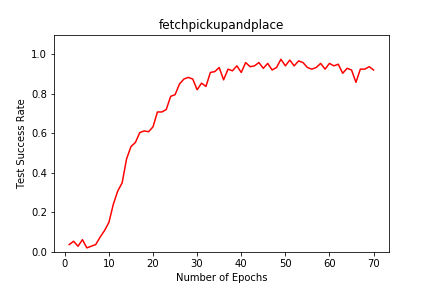}
    \end{minipage}
    \begin{minipage}{0.4\textwidth}
    \centering
    \includegraphics[width=\textwidth]{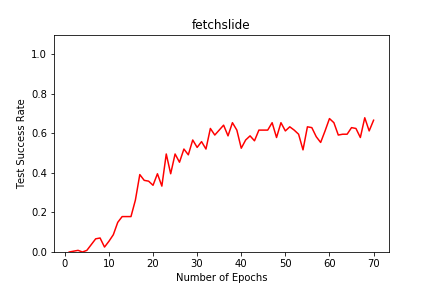}
    \end{minipage}
    \caption{Fetch Environments}
\end{figure}

\begin{figure}[h]
    \centering
    \begin{minipage}{0.4\textwidth}
    \centering
    \includegraphics[width=\textwidth]{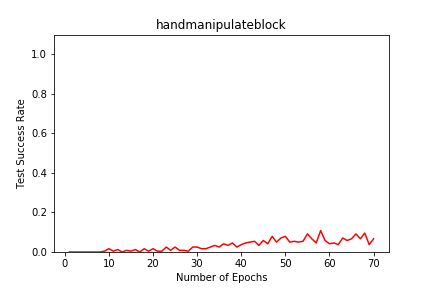}
    \end{minipage}
    \begin{minipage}{0.4\textwidth}
    \centering
    \includegraphics[width=\textwidth]{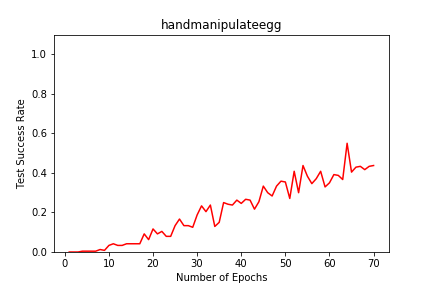}
    \end{minipage}
    \centering
    \begin{minipage}{0.4\textwidth}
    \centering
    \includegraphics[width=\textwidth]{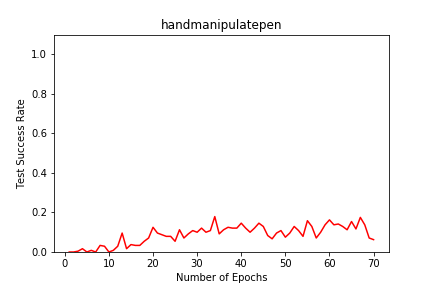}
    \end{minipage}
    \begin{minipage}{0.4\textwidth}
    \centering
    \includegraphics[width=\textwidth]{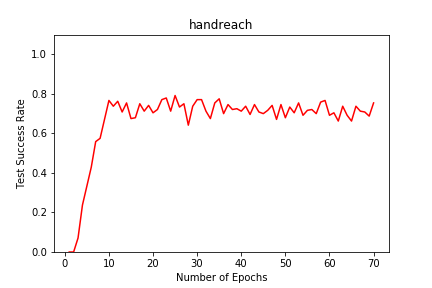}
    \end{minipage}
    \caption{Hand Environments}
\end{figure}

None of the hand environments reached the optimal performance even when run for $70$ epochs. Barring \textit{handmanipulateegg}, the other environments had reached saturation. There is scope for future work on improving performance.

To put things into perspective, \cite{HER} report in a blog that \textit{handmanipulateblock} reaches a success rate of $0.5$ in just 25 epochs with their machines. We observed saturation at around $0.2$. Just a reduction in $7$ CPUs does not seem to be the right reasoning for this discrepancy. The authors also do not mention the hyperparameters for the environments.
\newpage



\section{Implementation of Rank Based Prioritized Experience Replay}

We use an implementation from \href{https://github.com/Damcy/prioritized-experience-replay}{here}. There is no principled reason to choose rank based prioritization over proportional prioritization other than the fact that rank based is more robust to small changes in TD error and proportional prioritization takes into account the magnitude of the TD error.

Each transition is given a rank between $[1,n]$ and the probability of picking transition with a rank $i$ is $\propto \frac{1}{i}$. An ideal implementation would use a sorted array, but since maintaining a sorted array is very expensive a Binary Heap is used as an approximation.

Sampling is done by using \href{https://en.wikipedia.org/wiki/Inverse_transform_sampling}{inverse transform sampling}. If a batch size of $b$ is required, the distribution is used and transitions are sampled $b$ times. However, to make sure the sampling is stratified, the numbers from $[1,n]$ are divided into $k$ buckets, where $k$ is usually the \textit{batch\_size} and each bucket is defined to have equal probability. Sampling once from each bucket now gives a stratified sample.

There seems to be only $1$ open source implementation for tensorflow-python for rank based and it turned out to be very slow for our purposes. A modified version of the code will be made available soon which gives $\approx 10^{3}$ speed with only added amortized costs. This speed up was obtained by building cumulative distributions that are required incrementally rather than all at once at the starting.

\section{TRPO Version of Hindsight Policy Gradients}
As mentioned in section 5, Hindsight Policy Gradients can be extended by to TRPO. In this section, we derive the TRPO version of the Hindsight Policy Gradients by following the same mathematical steps as in \cite{TRPO}. Consider a trajectory denoted by $\tau$, obtained when an agent tries to achieve a goal $g'$. The performance of a policy is defines as: 
\begin{equation}
\eta(\pi) = \E_{g\sim G, \tau|\pi,g}\left(\sum_{t=0}^{\infty} \gamma^t r_t \right)
\end{equation}
Assume that the agent was pursuing a goal $g'$. Consider $\pi_g$ as a goal conditioned policy for pursuing a goal $g$. Let $\pi'$ be the old policy while $\pi$ is the current policy. Using the definition of value function and advantage function, we arrive at
\begin{equation}
\eta(\pi) = \eta(\pi'_g) + \E_{g\sim G, \tau|\pi,g}\left(\sum_{t=0}^{\infty} \gamma^t A^{\pi'_g}(s_t, a_t| g) \right)
\end{equation}
This can be simplified using the steady state probability distribution as
\begin{equation}
\eta(\pi) = + \eta(\pi'_{g}) + \E_{g\sim G, s\sim\rho_{\pi_g}, a\sim\pi_g} \left( A^{\pi'_g}(s,a) \right)
\end{equation}
A surrogate objective function $L(\pi)$ is defined as
\begin{equation}
L(\pi) = \E_{g\sim G, s\sim\rho_{\pi'_{g'}}, a\sim\pi_g} \left( A^{\pi'_g}(s,a) \right)
\end{equation}
\begin{equation}
L(\pi) = \E_{g\sim G, s\sim\rho_{\pi'_{g'}}, a\sim\pi'_{g'}} \left( \frac{\pi_g(a|s)}{\pi'_{g'}(a|s)} A^{\pi'_g}(s,a) \right)
\end{equation}
It can be verified that at the first order, this objective gives the same policy gradient at $\pi = \pi'$. The monotonic improvement is yet to be proven for this surrogate objective function. The way to prove it is to show a bound on the following quantity
\begin{equation}
\E_{g\sim G}\left[\E_{s\sim\rho_{\pi'_{g'}}, a\sim\pi_g} \left( A^{\pi'_g}(s,a) \right) - \E_{s\sim\rho_{\pi_g}, a\sim\pi_g} \left( A^{\pi'_g}(s,a) \right)\right]
\end{equation}

\section{Effect of alternate goal sampling on performance}
Alternate goals are sampled at storage time for appending the transitions with these alternate goals. As discussed before, a uniform alternate goal sampling allows for multiple copies of the transitions to be stored in the buffer. This makes the agent sample these transitions more often than necessary simply because updating the priority of one of the multiple copies doesn't change the priority of all of them. Hence this transition will be sample unnecessarily large number of times. This effect can be seen in play in the following performance curves

\begin{figure}[h]
    \centering
    \begin{minipage}{0.4\textwidth}
    \centering
    \includegraphics[width=\textwidth]{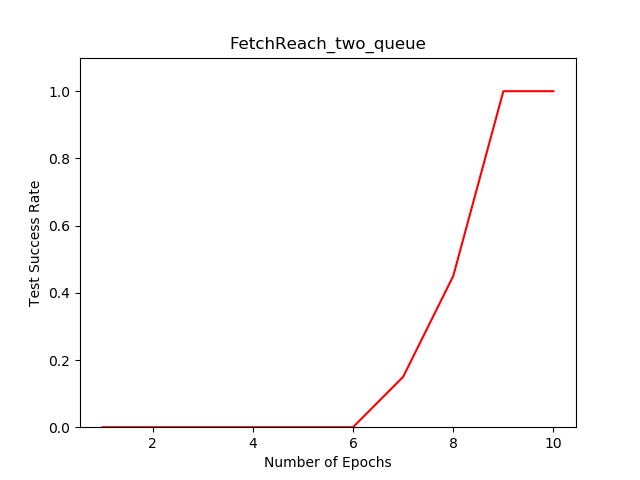}
    \end{minipage}
    \begin{minipage}{0.4\textwidth}
    \centering
    \includegraphics[width=\textwidth]{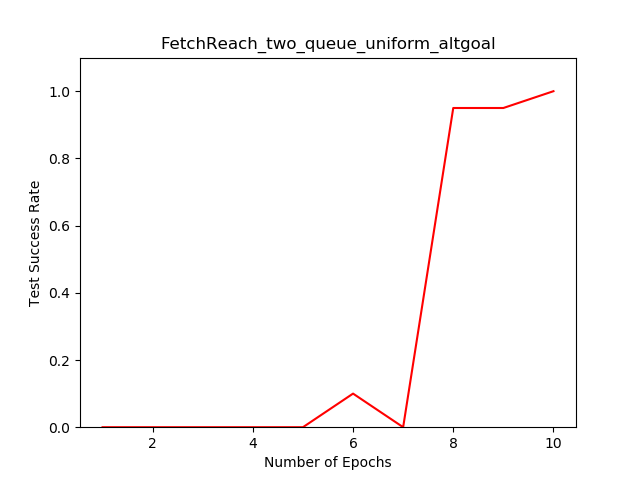}
    \end{minipage}
    \caption{The right hand side is an agent learning from samples appended with alternate goals sampled uniformly from future while, the left hand side is an agent learning from samples appended with alternate goals as mentioned in section $3.2$. As it can be seen, it takes longer for the agent to reach a success rate of $1.0$ when, uniform sampling of alternate goals is used for appending to transition at storage time. This fact has been established over multiple environments at various hyper-parameter values.}
\end{figure} 
\section{Comparing \texttt{single\_queue} and \texttt{two\_queues}}
As mentioned before, \texttt{single\_queue} gives the agent the degree of freedom to pick it's own $replay\_k$ i.e., the ratio of alternate goal appended samples to the actual goal appended samples. This degree of freedom improves the performance of \texttt{single\_queue} as transitions are sampled for replay based solely on the TD error whereas in \texttt{two\_queues}, there is a constraint on the $replay\_k$ which doesn't allow such a sampling. The performance curves for both \texttt{single\_queue} and \texttt{two\_queues} are as follows

\begin{figure}[h]
	\centering
    \begin{minipage}{0.4\textwidth}
    \centering
    \includegraphics[width=\textwidth]{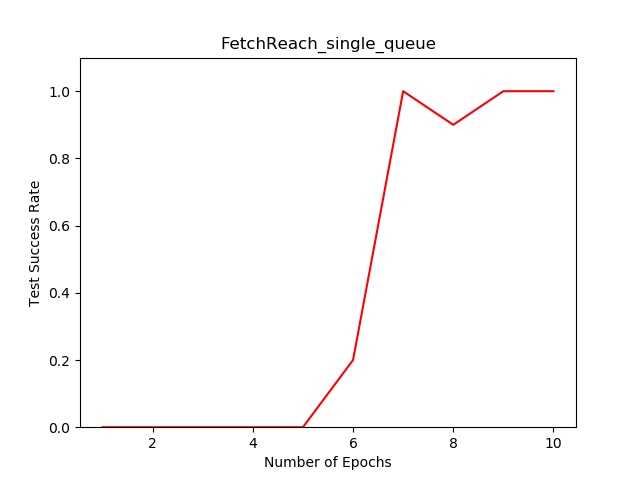}
    \end{minipage}
    \begin{minipage}{0.4\textwidth}
    \centering
    \includegraphics[width=\textwidth]{FetchReach_two_queue_plot.png}
    \end{minipage}
    \caption{The right hand side is an agent learning using \texttt{single\_queue} strategy while, the left hand side is an agent learning using \texttt{two\_queues} strategy. As it can be seen, it takes longer for the agent to reach a success rate of $1.0$ when it uses \texttt{two\_queues} strategy. This again has been observed over multiple experiments.}
\end{figure}



\end{appendix}

\end{document}